\documentclass[journal]{IEEEtran}
\usepackage{cite}

\ifCLASSINFOpdf
   \usepackage[pdftex]{graphicx}
\else
\fi
\usepackage{multirow}
\usepackage{esdiff}
\usepackage{subfigure}
\usepackage{adjustbox}
\usepackage{amsmath}
\usepackage{bbold}
\usepackage{bm}
\begin{document}
%
\title{Unsupervised AER Object Recognition  Based on Multiscale Spatio-Temporal Features and Spiking Neurons\\}
%
%
%

\author{Qianhui~Liu,~
	Gang~Pan,~
        Haibo~Ruan,~
        Dong~Xing,~
        Qi~Xu,~
        and~Huajin~Tang~
\thanks{Q. Liu, H. Ruan, D. Xing, Q. Xu, and H. Tang are with College of Computer Science, Zhejiang University, Hangzhou 310027, China. (e-mail: qianhuiliu@zju.edu.cn; hbruan@zju.edu.cn; dongxing@zju.edu.cn; xuqi123@zju.edu.cn;  huajin.tang@gmail.com).}
\thanks{G. Pan is with the State Key Lab of CAD\&CG, Zhejiang University, Hangzhou 310058, China. (e-mail: gpan@zju.edu.cn). (Corresponding author: Gang Pan)}
}

\maketitle

\begin{abstract}
 This paper proposes an unsupervised address event representation (AER) object recognition approach. The proposed approach consists of a novel multiscale spatio-temporal feature (MuST) representation of input AER events and a spiking neural network (SNN) using spike-timing-dependent plasticity (STDP) for object recognition with MuST. MuST extracts the features contained in both the spatial and temporal information of AER event flow, and meanwhile forms an informative and compact feature spike representation. We show not only how MuST exploits spikes to convey information more effectively, but also how it benefits the recognition using SNN. The recognition process is performed in an unsupervised manner, which does not need to specify the desired status of every single neuron of SNN, and thus can be flexibly applied in real-world recognition tasks. The experiments are performed on five AER datasets including a new one named GESTURE-DVS. Extensive experimental results show the effectiveness and advantages of this proposed approach.
\end{abstract}


%
\IEEEpeerreviewmaketitle

\section{Introduction}
%
%
%
%
%
%

%
%

\IEEEPARstart
{N}{euromorphic} engineering takes inspiration from biology in order to construct brain-like intelligent systems and has been applied in many fields such as pattern recognition, neuroscience, and computer vision \cite{indiveri2015memory,monroe2014neuromorphic}. Address event representation (AER) sensors are neuromorphic devices imitating the mechanism of human retina. Traditional cameras usually record the visual input as images at a fixed frame rate, which would suffer from severe data redundancy due to the strong spatio-temporal correlation of the scene. This problem could be solved to a large extent with AER vision sensors, which naturally respond to moving objects and ignore static redundant information. Each pixel in the AER sensor individually monitors the relative changes of light intensity of its receptive field. If the change exceeds a predefined threshold, an event will be emitted by that pixel. Each event carries the information of timestamp (the time when the event was emitted), address (the position of the corresponding pixel in the sensor) and polarity (the direction of the light change, i.e., dark-to-light or light-to-dark). The final output of the sensor is a stream of events collected from each pixel, encapsulating only the dynamic information of the visual input. 
Compared with traditional cameras, AER sensors have the advantage of maintaining an asynchronous, high-temporal-resolution and sparse representation of the scene. Commonly used AER sensors include the asynchronous time-based image sensor (ATIS) \cite{posch2011qvga}, dynamic vision sensor (DVS) \cite{lichtsteiner2008128,lenero20113}, dynamic and active pixel vision sensor (DAVIS) \cite{brandli2014240}.

The output of AER vision sensor is event-based; however, there remain open challenges on how to extract the features of events and then to design an appropriate recognition mechanism.
Peng et al.\cite{peng2017bag} proposed a feature extraction model for AER events called Bag of Events (BOE) based on the joint probability distribution of events. 
In addition, there are some existing works inspired by the cortical mechanisms of human vision, with a hierarchical organization that can provide features of increasing complexity and invariance to size and position \cite{serre2007feedforward}. 
 Chen et al.\cite{chen2012efficient} proposed an algorithm to extract size and position invariant line features for recognition of objects, especially human postures in real-time video sequences from address-event temporal-difference image sensors. 
Zhao et al. \cite {zhao2015feedforward} presented an event-driven convolution-based network for feature extraction that takes data from temporal contrast AER events, and also introduced a forgetting mechanism in feature extraction to retain timing information of events into features. 
Lagorce et al. \cite{lagorce2017hots} proposed the HOTS model, which relies on a hierarchical time-oriented approach to extract spatio-temporal features called time-surfaces from the asynchronously acquired dynamics of a visual scene.  
The time-surfaces are using relative timings of events to give contextual information.
Orchard et al. \cite{orchard2015hfirst} proposed the HFirst model, in which a spiking hierarchical model with four layers was introduced for feature extraction by utilizing the timing information inherently presented in AER data. 
\begin{figure*}[!t]
	
	\centering
		\includegraphics[width=0.88\textwidth]{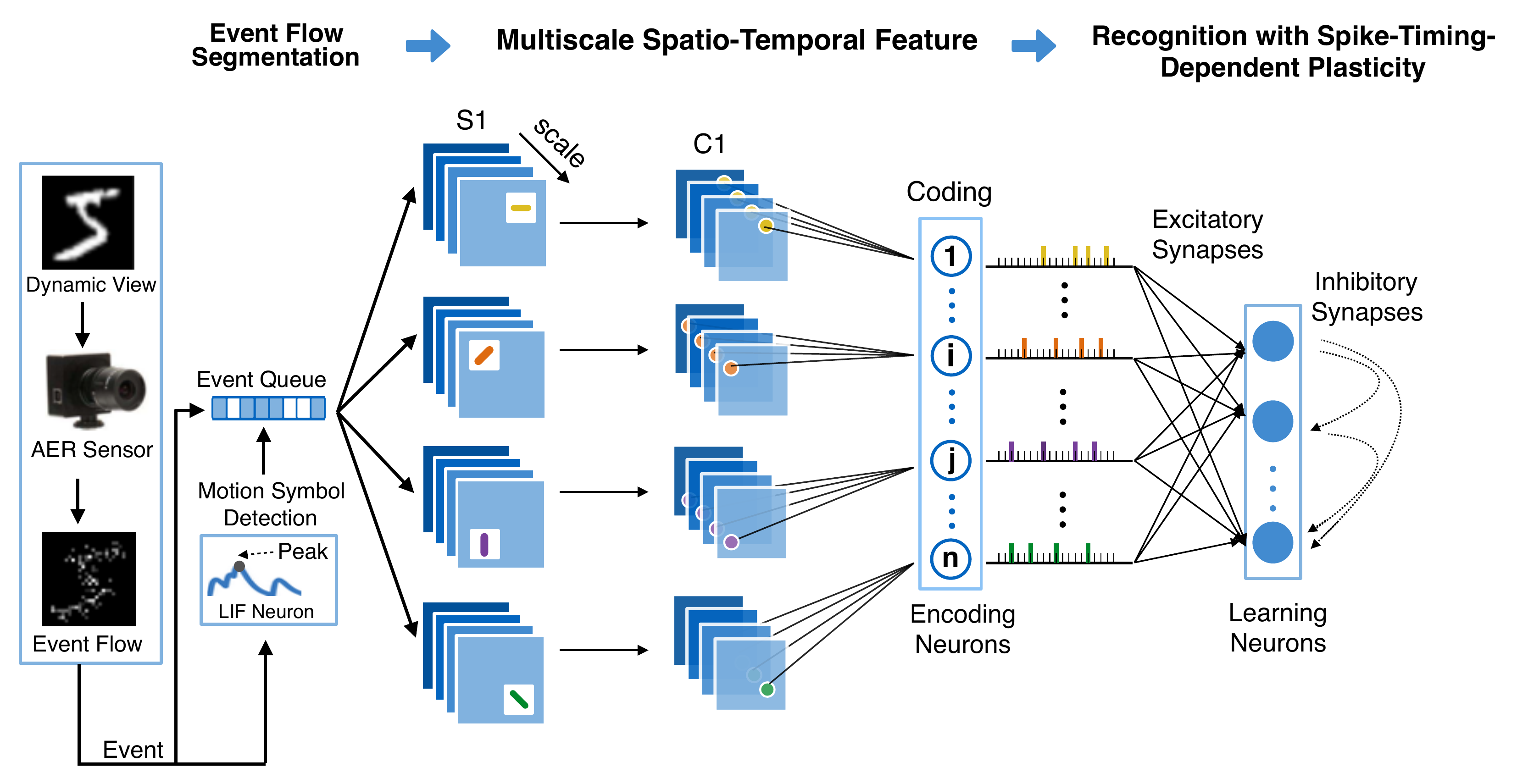} %
	\caption{The flow chart of the proposed AER object recognition. 
The event flow from AER sensor are sent concurrently to motion symbol detection (MSD) \cite{zhao2015feedforward} and event queue. 
MSD adaptively partitions the events waiting to be sent in event queue into segments, and streams the events segment by segment to neurons in $S1$ layer for spatio-temporal feature extraction. Neurons have their own scale of receptive field and respond best to a certain orientation. The neuron responses reflect the strength of features, which cover both the spatial features of different scales and orientations and temporal information. Neurons of the same receptive scale and orientation are organized into one feature map (denoted by blue squares) and the max responses in adjacent non-overlapping 2 $\times$ 2 neuron regions of each feature map reach the $C1$ layer. 
The $C1$ features are coded to spikes and multiscale features having the same orientation and position in $C1$ maps flow into the same encoding neuron.  
The encoding neurons emit spikes to trigger learning neurons and the relative timing of spikes will trigger the spike-timing-dependent plasticity (STDP) on excitatory synapses during training. Each learning neuron inhibits others through inhibitory synapses (denoted by dash lines), ensuring different neurons learn different patterns. After training, each learning neuron will be assigned a class label based on its sensitivity to patterns of different classes. The final recognition decision is determined by averaging the firing rates of learning neurons per class and choosing the class with the highest average firing rate.
}
	
	\label{fig:framework}
	
\end{figure*}

In addition, biological study of the visual ventral pathway indicates that vision sensing and object recognition in the brain are performed in the form of spikes \cite{memmesheimer2014learning}. 
Several coding hypotheses \cite{panzeri2010sensory,hu2016brain} have been proposed from different aspects to explain how these spikes represent information in the brain. Neurons in the visual cortex have been observed to precisely respond to the stimulus on a millisecond timescale \cite{butts2007temporal}. This supports the hypothesis of temporal coding, which considers that information about the stimulus is contained in the specific precise spike timing of the neuron. To implement the temporal coding,  we need to specify the coding function to map the features of AER events to precise spike timings. 
How to select a coding function that can better convey the information contained in features into spikes and contribute to the object recognition becomes a key question.
We also design the coding mechanism from spatial perspective since the spatial information of feature spikes also takes effects in object recognition.

Inspired by previous works, we introduce an encoding scheme for AER events that extracts the spatio-temporal features of raw events and forms a feature spike representation.
Considering that biological neurons are inherently capable of processing temporal information, we present a cortex-like hierarchical feature extraction based on leaky integrate-and-fire (LIF) spiking neurons with spatial sensitivity.
The responses of these neurons are accumulated along the time axis and reflect the strength of current spatio-temporal features.
We also propose the coding mechanism to obtain the spatio-temporal feature spikes, which consists of the natural logarithmic temporal coding function and multiscale spatial fusion.
Through the proposed coding function, we obtain the feature spikes with even temporal distribution. We will show that these spikes are more informative and contribute to the recognition using SNN. 
Meanwhile, the spatio-temporal features of multiple scales are highly correlated and are fused to spike-trains to form a multiscale spatio-temporal feature representation, which we have called MuST.

Since MuST is in the form of spikes, it is natural to employ the spiking neural network (SNN) to learn the spike patterns.
 Compared with traditional classifiers, SNNs are more natural to interpret the information processing mechanisms of the brain \cite{yu2013rapid}, 
 and more powerful on processing both spatial and temporal information \cite{zhang2018plasticity}. 
 In addition, SNNs have the advantage of low power consumption, for example, current implementations of SNN on neuromorphic hardware use only a few $nJ$ or even $pJ$ for transmitting a spike \cite{diehl2015unsupervised}. 
Most existing works for AER object recognition, such as  \cite{zhao2015feedforward} and \cite{ma2017event}, have chosen supervised classifiers of SNN for recognition. 
These supervised classifiers need to specify the desired status of firing or not or even the firing time of neurons. However, setting the desired status to every single neuron is intricate and tedious in real-world recognition tasks.
We consider the unsupervised learning rule spike-timing-dependent plasticity (STDP)\cite{bi1998synaptic} of SNN. STDP works by considering the relative timing of presynaptic and postsynaptic spikes. According to this rule, if the presynaptic neuron fires earlier (later) than the postsynaptic neuron, the synaptic weight will be strengthened (weakened). Through the STDP learning, each postsynaptic neuron naturally becomes sensitive to one or some similar presynaptic spike patterns.
There are some existing works that have shown the powerful ability of STDP to learn the spike patterns. 
Diehl et al. \cite{diehl2015unsupervised} proposed a SNN for image recognition that employs STDP learning to process the Poisson-distributed spike-trains with firing rates proportional to the intensity of the image pixel.
 Iyer et al. \cite{iyer2017unsupervised}  applied the Diehl's model \cite{diehl2015unsupervised} on native AER data. Experiments on the N-MNIST dataset \cite{orchard2015converting} show that the method provides an effective unsupervised application on AER event streams.  
 Zheng et al.\cite{zheng2018sparse} presented a spiking neural system that uses STDP-based HMAX to extract the spatio-temporal information from spikes patterns of the convolved image. 
Panda et al. \cite{panda2016unsupervised} presented a regenerative model that learns the hierarchical feature maps layer-by-layer in a deep convolutional network using STDP. 
Our major contributions can be summarized as follows:
\begin{itemize}
	\item We propose an unsupervised recognition approach for AER object, which performs the task using MuST for encoding the AER events and STDP learning rule of SNN for object recognition with MuST. This approach does not require a teaching signal or setting the desired status of neurons in advance, and thus can be flexibly applied in real-world recognition tasks.
	\item 

We present MuST which not only exploits the information contained in the input AER events, but also forms a new representation that is suitable for the recognition mechanism. MuST extracts the spatio-temporal features of AER events based on LIF neuron model, and forms a feature spike representation which consumes less computational resources while still maintaining comparable performance.
	\item Extensive experimental analysis shows that our recognition approach, processed in an unsupervised way, can achieve comparable performance to existing, supervised solutions. 
\end{itemize}
The rest of this paper is organized as follows. Section \ref{sec:overview} overviews the flow of information processing in this approach. Section \ref{sec:feature}-\ref{sec:stdp} describes the details of this recognition approach. The experimental results are explained in Section \ref{sec:result}. In section \ref{sec:conclu}, we come to our conclusion.

\section{Overview of the proposed approach}\label{sec:overview}
　The proposed AER object recognition consists of three parts, namely the Event Flow Segmentation, Multiscale Spatio-Temporal Feature (MuST) and Recognition with STDP, as shown in Fig. \ref{fig:framework}. We will overview the flow of information processing in this approach as follows.

\textbf{Event Flow Segmentation:}
Our object recognition approach is driven by raw events from the AER sensor. However, 
 it is still a daunting task to explore how to use each single event as a source of meaningful information \cite{peng2017bag}.
In addition, due to the high temporal resolution of the sensor, the time intervals between two successive events can be very small (100$ns$ or less). For the efficiency of computation and energy use, 
existing works \cite{zhao2015feedforward, peng2017bag} heuristically partition events into multiple segments and then perform the feature extraction and recognition based on these segments. 
We maintain an event queue to store the input events waiting to be sent to the next layer, and meanwhile apply the motion symbol detection (MSD) \cite{zhao2015feedforward} to adaptively partition the events according to their statistical characteristics, which is more  flexible than the partition methods based on fixed time slices or fixed event numbers.
Events from the AER sensor are sent concurrently to the event queue and MSD. MSD consists of 
  a leaky integrate-and-fire (LIF) neuron and a peak detection unit. The neuron receives the stimuli of events and then updates its total potential. The peak detection is applied to locate temporal peaks on the neuron's total potential. A peak is detected when many events have occurred intensively, which indicates enough information has been gathered. Therefore, once the peak is detected, events in the event queue emitted before the peak time will be sent as a segment to the next part. 

\textbf{Multiscale Spatio-Temporal Feature (MuST):}
The events are sent to the $S1$ layer, which consists of neurons having their own scale of receptive field and responding best to a certain orientation. $S1$ neurons accumulate the responses which reflect the strength of spatial features. The timing information of events is also recorded in the responses of $S1$ neurons
because of the spontaneous leakage of neurons.
Each neuron associates to one pixel in the sensor and neurons of the same receptive scale and orientation are organized into one feature map.
 Each feature map in $S1$ is divided into adjacent non-overlapping 2 $\times$ 2 neuron regions and the max neuron responses in each region reach the $C1$ layer.  
The neuron responses (features) in $C1$ layer are coded into the form of spikes for recognition.
The strength of the feature is in a logarithm manner mapped to the timing of spike by temporal coding, and multiscale features having the same orientation and position in $C1$ feature maps are fused as a spike-train flowing to one encoding neuron, forming the MuST representation of AER events for recognition.

\textbf{Recognition with STDP:} 
The encoding neurons emit spikes to excite the learning neurons of the SNN. According to STDP, the relative timing of spikes of the presynaptic encoding neuron and postsynaptic learning neuron triggers the synaptic weight adjustment during training. The spikes from one learning neuron also inhibit the other learning neurons. This lateral inhibition prevents neurons from learning the same MuST pattern.
After training, each learning neuron will be assigned a class label based on its sensitivity to patterns of different classes. The final recognition decision for an input pattern is determined by averaging the firing rates of learning neurons per class and choosing the class with the highest average firing rate.

\section{Multiscale Spatio-Temporal Feature}
\label{sec:feature}

The current theory of the cortical mechanism  has been pointing to a hierarchical and mainly feedforward organization \cite{zhao2015feedforward}. In the primary visual cortex (V1), two classes of functional cells -- simple cells and complex cells are founded \cite{hubel1962receptive}. Simple cells respond best to stimuli at a particular orientation, position and phase within their relatively small receptive fields. Complex cells tend to have larger receptive fields and exhibit some tolerance with respect to the exact position within their receptive fields. Further, plasticity and learning certainly occur at the level of inferotemporal (IT) cortex and prefrontal cortex (PFC), the top-most layers of the hierarchy\cite{serre2007robust}.

Inspired by the visual processing  in the cortex, we introduce the following mechanisms in our recognition approach:
1) We model the object recognition a hierarchy of $S1$ layer, $C1$ layer, encoding layer and learning layer. 
2) MuST feature extraction consists of $S1$ and $C1$ layer, composed of simple cells and complex cells respectively. Simple cells combine the input with a bell-shaped tuning function to increase feature selectivity and complex cells perform the max pooling operation to increase feature invariance. We use LIF neurons to model the simple and complex cells.  
The LIF model has been used widely to simulate biological neurons and is inherently good at processing temporal information. The responses of the neurons just reflect the strength of spatio-temporal features with selectivity and invariance. We also propose a coding mechanism from the temporal and spatial perspectives, aiming to form a feature spike representation to better exploit the information in raw events for recognition.
3) The STDP rule  models the learning at the high layer of the  hierarchy and learns sophisticated features of objects, which will be described in detail in the next section. 
 
 In this section, we will propose the multiscale spatio-temporal feature representation of the raw AER events. 
\subsection{Spatio-Temporal Feature Extraction}
We conduct the feature extraction using bio-inspired hierarchical network composed of LIF neurons with a certain receptive scale and orientation, which takes into account both the  temporal and spatial information encapsulated in AER events.
This network contains two layers named $S1$ layer and $C1$ layer, mimicking the simple and complex cells in primary visual cortex $V1$ respectively.
An event-driven convolution is introduced in neurons of the $S1$ layer,
 and a max-pooling operation is used in the $C1$ layer.

\subsubsection{$S1$ layer}
Each event in the segment is sent to the $S1$ layer, in which the input event is convolved with a group of Gabor filters \cite{zhao2015feedforward}. The function of Gabor filter can be described with the following equation:
  \begin{equation}\label{Gabor1}
  G(\Delta x,\Delta y;\sigma,\lambda,\theta) 
  = \exp(-\frac{X^2+\gamma ^2 Y^2}{2\sigma ^2})\cos(\frac{2\pi}{\lambda}X)
  \end{equation}
  \begin{equation}\label{Gabor2}
  X = \Delta x \cos\theta + \Delta y\sin\theta 
  \end{equation}
  \begin{equation}\label{Gabor3}
  Y = -\Delta x\sin\theta + \Delta y\cos\theta 
  \end{equation}
 where  $\Delta x$ and $\Delta y$ are the spatial offsets between the pixel position ($x$,$y$) and the event address ($e_x$, $e_y$), $\gamma$ is the aspect ratio.  The wavelength $\lambda$ and effective width $\sigma$ are parameters determined by scale $s$. 
Each filter models a neuron cell that has a certain scale $s$ of receptive field and responds best to a certain orientation $\theta$. 
Each neuron associates to one pixel in the sensor and neurons of the same receptive scale and orientation are organized into one feature map.
The responses of neurons in feature maps are initialized as zeros, then updated by accumulating each element of the filters to the maps at the position specified by the address of each event. The response of the neuron at position $(x,y)$ and time $t$ in the map of specific scale $s$ and orientation $\theta$ can be described as:
 \begin{multline}\label{ResMap}
r(x,y,t;s,\theta) = 		
\sum_{e \in E(t)} 
\mathbb{1}\{x \in \mathcal{X}(e_x)\}\mathbb{1}\{y \in \mathcal{Y}(e_y)\}\\
\exp(-\frac{t-e_t}{\tau_{leak}})G(x-e_x,y-e_y;\sigma(s),\lambda(s),\theta)
\end{multline}
 where $E(t)$ denotes the set of events which are emitted before the time $t$ in the current segment, $\mathbb{1}\{.\}$ is the indicator function,  $\mathcal{X}(e_x) = [e_x -s, e_x + s]$ and  $\mathcal{Y}(e_y) = [e_y -s, e_y + s]$ denote the receptive field of the neuron, and $\tau_{leak}$ denotes the decay time constant.
 Since the parameters $\sigma$ and $\lambda$ in function $G$ are determined by $s$, we herein use  $\sigma(s)$ and $\lambda(s)$ instead.
This computation process can also be explained in another way. 
When the address of the current event $e$ is in the receptive fields of the neuron, the response $r(x,y,t;s,\theta)$ of the neuron is increased by $G(\Delta x,\Delta y; s,\theta)$. Otherwise, the neuron response keeps decaying exponentially. The decay dynamics of the response are:
 \begin{equation}\label{conductance}
 \tau_{leak}\diff{r(x,y,t;s,\theta)}{t} = - r(x,y,t;s,\theta)
 \end{equation}
 With the exponential decay, the impact of earlier events is reduced on the current responses, and the precise timing information of each event can be captured in the responses. 
 
 \subsubsection{$C$1 layer}
 Each feature map in $S1$ layer is divided into adjacent non-overlapping $2 \times 2$ cell regions, namely $S1$ units. The responses of $C1$ cells are obtained by max pooling over the responses in $S1$ units. The pooling operation causes the competition among $S1$ cells inside a unit, and  high-response features (considered as representative features) will reach the $C1$ maps.
  After the pooling operation, the number of cells in $C1$ maps is $1/4$ of that in $S1$ maps. This pooling operation decreases the number of required neurons in latter layers and makes the features locally invariant to size and position. 
 \begin{figure}[t]
 	\centering
 	\includegraphics[width=0.75\linewidth, bb = 34 12 500 340]{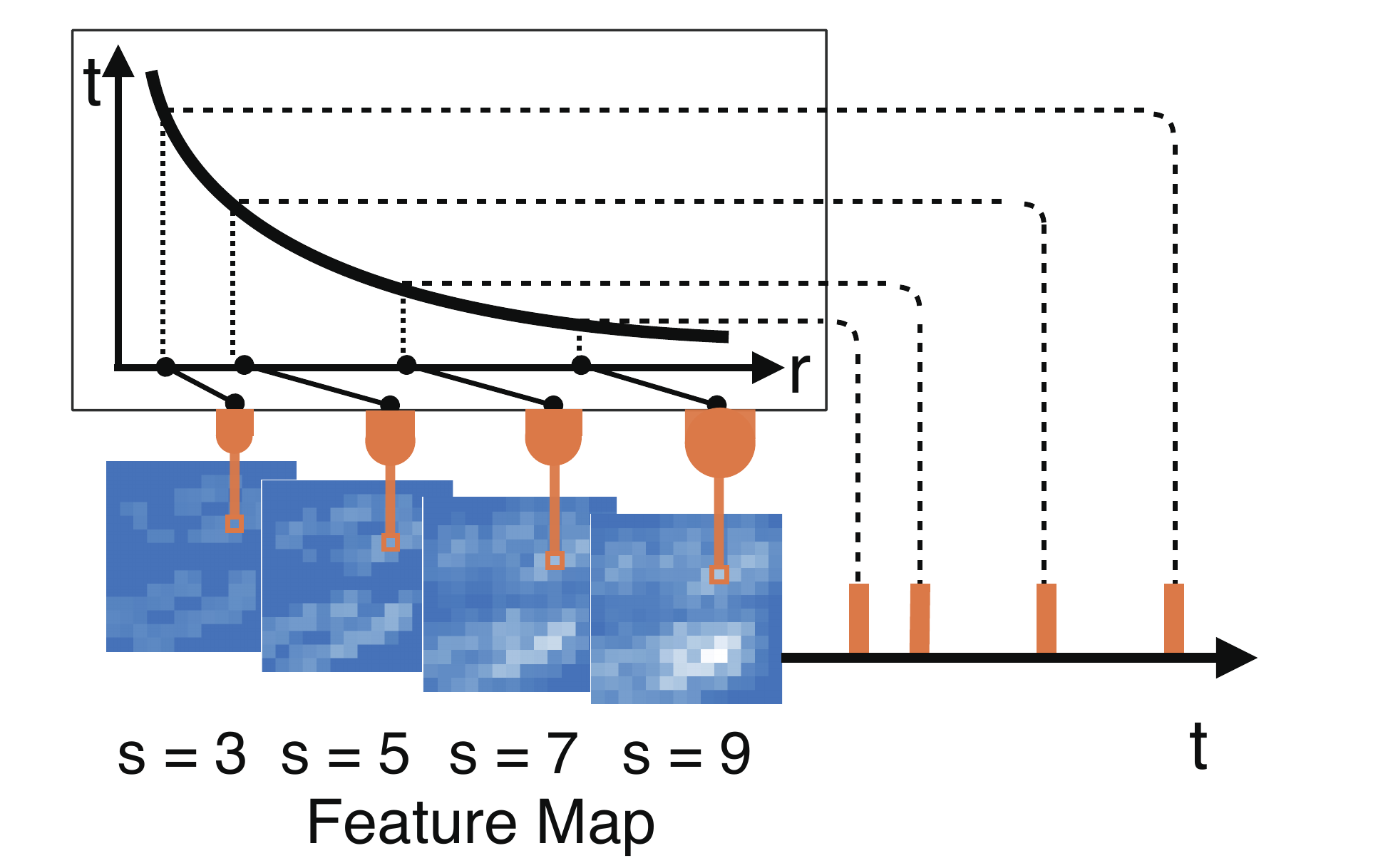}
 	\caption{Illustration of coding mechanism. Four blue squares denote the C1 feature maps of four different scales having the orientation of $45^{\circ}$. Four responses having same position in these four feature maps are chosen for illustration. These responses are converted to spikes by the logarithm coding function and then be fused into a spike-train. The lighter a pixel looks in the feature map, the higher is its response value, and the earlier is its corresponding spike timing. }
 	\label{fig:MuSTf}
 \end{figure}
  \begin{figure}[b]
 	\centering
 		\includegraphics[width=0.65\linewidth, bb = 19 10 470 380]{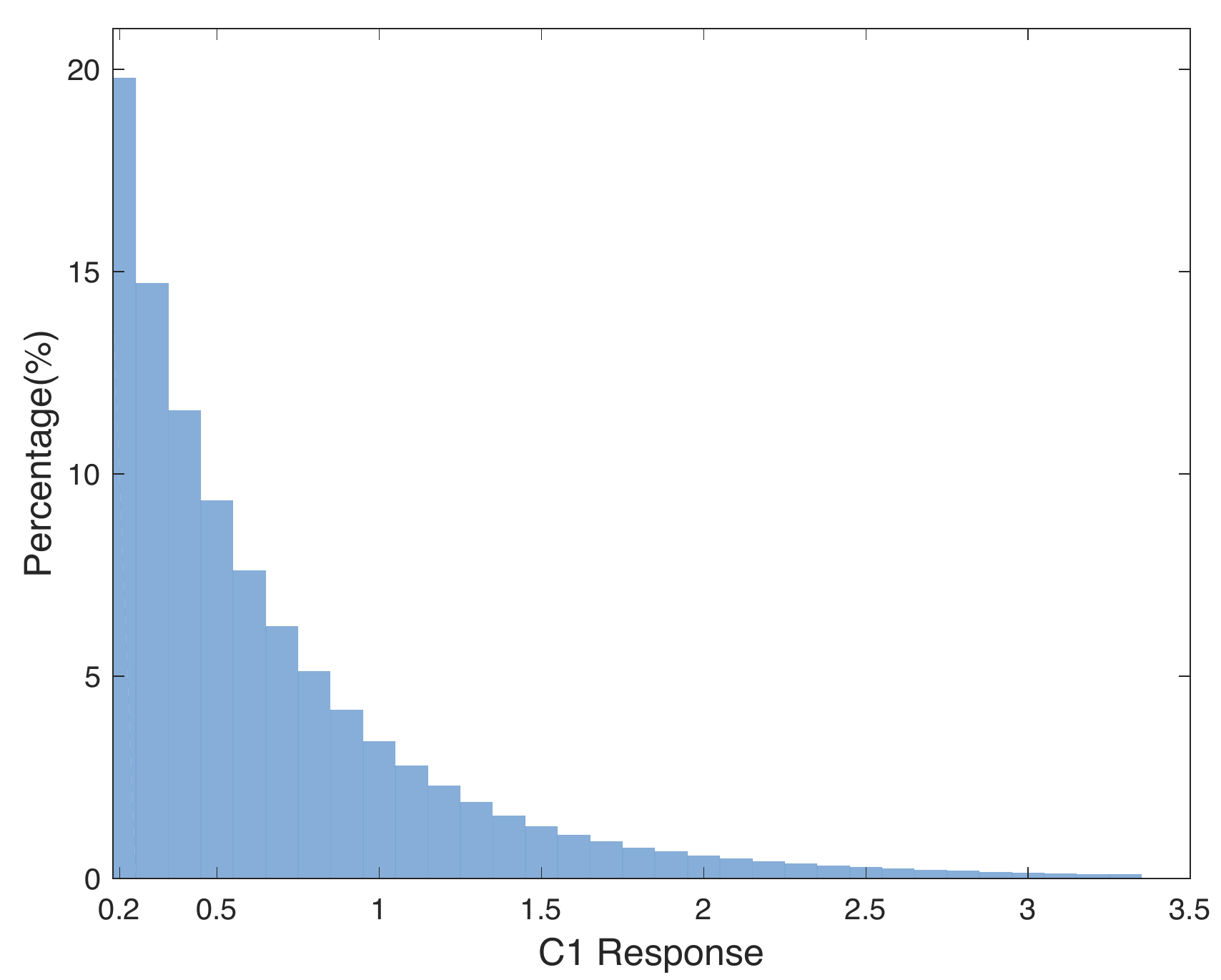}
 	\caption{The distribution of $C1$ feature responses on MNIST-DVS dataset. Each bin of the histogram has a nonoverlapping span of $0.1$. The height of each bin indicates the average proportion of the C1 responses in the corresponding span. We consider the features with responses smaller than $0.2$ noises and ignore them. }
 	\label{fig:c1Response}
 \end{figure}

\subsection{Coding to Spike-Trains}
The spatio-temporal features in $C1$ maps will be coded to spike-trains. A spike carries the information of its timestamp and address.
We propose a coding mechanism to convert the strength of feature to the spike timing by a natural logarithm function of temporal coding, and to map the position of feature to the address of spike by multiscale fusion.
This procedure is illustrated in Fig. \ref{fig:MuSTf}, and the details are described as follows.

The  feature responses in $C1$ maps are used to generate spike timings by latency coding scheme \cite{panzeri2010sensory,hu2016brain}.
Features with the maximum response values, which are considered to activate the spike more easily, correspond to the minimum latency and will fire early; features with smaller values will fire later or even not fire.

We focus on finding an appropriate coding function in order to fully utilize the information contained in features for the following recognition.
We randomly choose 1000 samples from MNIST-DVS dataset, and show the distribution of $C1$ responses in Fig. \ref{fig:c1Response}. It can be seen that the distribution of features is heavily skewed. 
Linear coding functions are used by many existing works \cite{liu2017fast,zhao2015feedforward} to convert these features to spikes for simplicity, but such functions cannot change the distribution of data and thus the temporal distribution of feature spikes are still skewed. This skewed distribution of feature spikes will lead to two problems: 1) higher-response features have less impact on recognition process.
It is because the distribution of feature spikes affects the recognition process.
The spikes of higher-response features are more sparsely distributed so that receptive neurons (learning neurons in our approach) are hard to accumulate responses high enough to emit spikes (because of the leakage of neurons). Therefore,
the information in these high-response features cannot be completely transmitted to the receptive neurons and cannot be fully utilized by the recognition process.
2) Considering that the information of features in SNN is contained in the timings of feature spikes, the features are considered similar if the timings of their spikes are close. Therefore, it is difficult to distinguish two features whose spikes are densely distributed in a short time window.


To solve these problems, feature responses in our approach are in a logarithm manner inversely mapped to spike timings.
For one specific feature response depicted as $r$ within the $C1$ layer, the corresponding spike timing $t_{spike}$ can be computed as follows:
\begin{equation}\label{logFunc}
t_{spike}= C(r) = u - v \ln (r)
\end{equation}
where $u$ and $v$ are normalizing factors ensuring that the spikes fire in the predefined time window $t_w$, $C$ denotes the coding function of response $r$. The settings of $u$ and $v$ are as follows:
$u = t_w \ln (r_{max}) / ( \ln (r_{max}) - \ln (r_{min}))$ and $v =  t_w / ( \ln (r_{max}) - \ln (r_{min}))$, where $r_{max}$ is the maximum feature response in the training set, $r_{min}$ is the user-defined minimum threshold, less than which the responses are set to be ignored. Section \ref{sec:result} will show the effects of this natural logarithm coding function.

We then attach the address information of features to Equation (\ref{logFunc}) and obtain Equation (\ref{equation:fusion}). The spikes that converted from  feature responses $r$ at the position $(x,y)$ in the feature maps are written as:
\begin{equation}\begin{aligned}
&t_{spike}= C(r | x,y,S,\Theta) = u - v \ln (r)\\
&s.t.\quad
r \in \left\{ r | r_x = x, r_y = y, r_s \in S, r_{\theta} \in \Theta \right\}
\end{aligned}
	\label{equation:fusion}
\end{equation}
where $r_s$ and $r_{\theta}$ denote the scale and orientation of $r$, $r_x$ and $r_y$ denote the position of $r$ in feature map, $S$ is a set of values of scale $s$, $\Theta$ is a set of values of orientation $\theta$. 

Unlike artificial neurons each of which represents information as a real value, a spiking neuron can convey multiple signals in the form of a spike-train, which is more flexible and more informative. Considering this characteristic of spiking neurons, certain features can be fused to make more efficient use of neurons and form a compact representation.
Inspired by \cite{serre2007robust} where features of neighboring scales are combined together, 
 in our implementation, feature spikes of multiple scales having the same position and orientation are fused to a spike-train, sharing the same spike address. That is, each encoding neuron is in charge of the conversion of multiscale $C1$ features. The spike-train that is converted from feature responses $r$ having position $(x,y)$ and orientation $\theta$ is comprised of a set of $t_{spike}$ in Equation (\ref{equation:fusion}), where $\Theta =\left\{ \theta \right\}$.
The following experiments in Section \ref{sec:result} will provide the analyses and effects of this multiscale fusion method.

\begin{figure}[t]
	\centering
		\includegraphics[width=\linewidth, bb = 10 15 770 320]{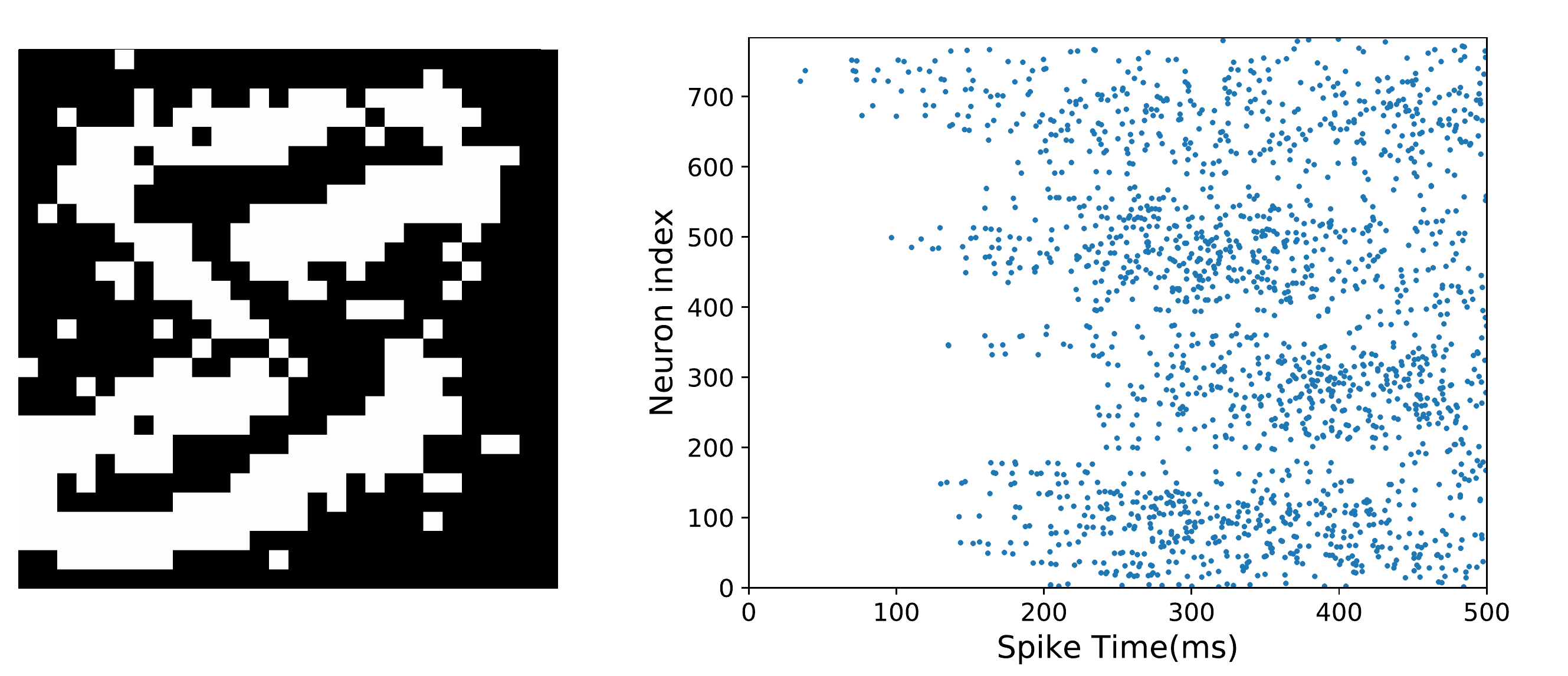}
	\caption{One reconstructed image and its MuST pattern. \textit{Left}: reconstructed image of the digit ``5" from the MNIST-DVS dataset. Black pixel denotes there is no event at this position and white pixel denotes there is at least one event at this position.  \textit{Right}: the corresponding MuST pattern.}
	\label{fig:spikePattern}
\end{figure}
Through this encoding scheme, each input segment has its own MuST representation. Fig. \ref{fig:spikePattern} shows a reconstructed image of an event segment in MNIST-DVS dataset \cite{lichtsteiner2008128}, and its corresponding MuST representation. 


\section{Recognition with spike-timing-dependent plasticity}\label{sec:stdp}
In this part, a network of spiking neurons (SNN) is developed to perform object recognition with MuST.
SNN simulates the fundamental mechanism of human brain and is good at processing spatio-temporal information.
STDP is used here as the unsupervised learning rule of SNN. Every neuron naturally becomes sensitive to one or some similar input spike patterns through STDP rather than approaching the desired status as in supervised learning. Due to the flexibility of STDP, it is more suitable for our real-world recognition tasks.
We will describe the network design and unsupervised learning method as follows.
\subsection{Network Design}
The input stimuli of this network are the MuST spike-trains the encoding neurons emit. Encoding neurons are fully connected to the learning neurons. These synaptic connections are excitatory and will be adjusted in training procedure. Each of the learning neurons inhibits all other ones by inhibitory synapses with the short delay $t_d$ and the weights of inhibitory synapses are set to the predefined value $w_{inh}$. This connectivity implements lateral inhibition. Once a learning neuron fires a spike, the inhibitory synapses transmit the stimuli to inhibit other learning neurons. 
The network design enables each neuron to represent one prototypical pattern or an average of some similar patterns, and  prevents a large number of neurons from representing only a few patterns.

During training, the weights of all excitatory synapses are firstly initialized with random values and are updated using STDP. When the training is finished, we assign a class to each neuron, based on its highest response to the different classes over one presentation of the training set. Only in this class assignment step are the labels being used. For the training of the network, we do not use any label information. During the testing phase, the predicted class for the input pattern is determined by averaging the firing rates of neurons per class and then choosing the class with the highest average firing rate.
\subsection{STDP Learning Rule}
STDP is a biological process that adjusts the weights of connections between neurons. 
Considering both the encoding and learning neurons emit multiple spikes, we employ the triplet STDP model \cite{pfister2006triplets} which is based on interactions of relative timing of three spikes (triplets).  Besides, triplet STDP has shown its computational advantage over standard STDP since it is sensitive for input patterns consisting of higher-order spatiotemporal spike pattern correlations \cite{gjorgjieva2011triplet}.

LIF model is chosen to describe the neural dynamics \cite{diehl2015unsupervised}. 
The membrane voltage $V$ of the neuron is described as:
\begin{equation}\label{voltage}
\tau \diff{V}{t} = V_{rest} - V+g_e(E_{exc} - V)+g_i(E_{inh} - V)
\end{equation}
where $\tau$  is the postsynaptic neuron membrane time constant, $V_{rest}$ the resting membrane potential, $E_{exc}$ and $E_{inh}$ the equilibrium potentials of excitatory and inhibitory synapses, and $g_e$ and $g_i$ the conductance variables of excitatory and inhibitory synapses, respectively. The conductance is increased by the synaptic weight $w$ at the time a presynaptic spike arrives, otherwise the conductance keeps decaying exponentially. If the synapse is excitatory, the decay dynamics of the conductance $g_e$ are:
\begin{equation}\label{conductance}
\tau_{ge}\diff{g_e}{t} = - g_e 
\end{equation}
where $\tau_{ge}$ is the time constant of an excitatory postsynaptic potential; if the synapse is inhibitory, $g_i$ is updated using the same equation but with the time constant of the inhibitory postsynaptic potential $\tau_{gi}$. 
When the neuron's membrane potential is higher than its threshold $V_{thr}$, the neuron will fire a spike and its membrane potential will be reset to $V_{reset}$. An adaptive membrane threshold \cite{zhang2003other,diehl2015unsupervised} is employed 
to prevent single learning neuron from dominating the response pattern. When the neuron fires a spike, the threshold $V_{thr}$ will be increased by ${V_{plus}}$. Otherwise the threshold $V_{thr}$  is described as:
\begin{equation}\label{adaptive}
\tau_{thr}\diff{V_{thr}}{t} = V_t - V_{thr} 
\end{equation}
where $V_t$ denotes the predefined membrane threshold. By incorporating such method, the more spikes a neuron fires, the higher its membrane threshold will be.

\begin{figure}[!t]
	\centering
	\includegraphics[width=1.\linewidth,bb=5 0 370 110]{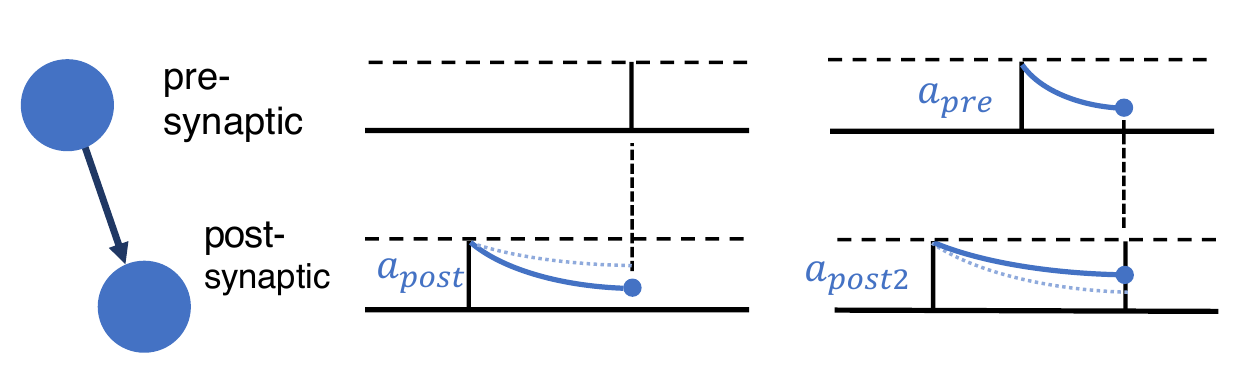}
	\caption{Two conditions of the triplet STDP rule. \textit{Left:} synaptic depression is induced using one postsynaptic trace when the presynaptic spike arrives. \textit{Right:} synaptic potentiation is induced using the post- and pre-synaptic traces when the postsynaptic spike arrives.}
	\label{fig:STDP}	
\end{figure}

The weight dynamics are computed using synaptic traces which model the recent spike history. Each synapse keeps tracks of one presynaptic trace $a_{pre}$ and two postsynaptic traces $a_{post}$ and $a_{post2}$. For simplicity, we use the Nearest-Spike interaction. As shown in Fig. \ref{fig:STDP}, every time a presynaptic spike arrives at the synapse, $a_{pre}$ is assigned to 1; otherwise $a_{pre}$ decays exponentially. The decay dynamic of the trace $a_{pre}$ is:
\begin{equation}\label{pretrace}
\tau_{a_{pre}}\diff{a_{pre}}{t} = - a_{pre}
\end{equation}
where $\tau_{a_{pre}}$ is the time constant of trace $a_{pre}$. The postsynaptic traces $a_{post}$ and $a_{post2}$ work the same way as the presynaptic trace but their assignments are triggered by a postsynaptic spike and they decay with the time constant $\tau_{a_{post}}$ and  $\tau_{a_{post2}}$ respectively.
When a presynaptic spike arrives at the synapse, the weight is updated based on the postsynaptic trace:
\begin{equation}\label{w_update}
\Delta w = A^- a_{post}
\end{equation}
where $A^-$ is the learning rate for presynaptic spike. When a postsynaptic spike arrives at the synapse the weight change $\Delta w$ is:
\begin{equation}\label{w_update2}
\Delta w = A^+ a_{pre} a_{post2}
\end{equation}
where $A^+$ is the learning rate.
%
\begin{figure*}[!t]
	\centering
	\includegraphics[width=0.7\textwidth,bb=100 20 6500 3200]{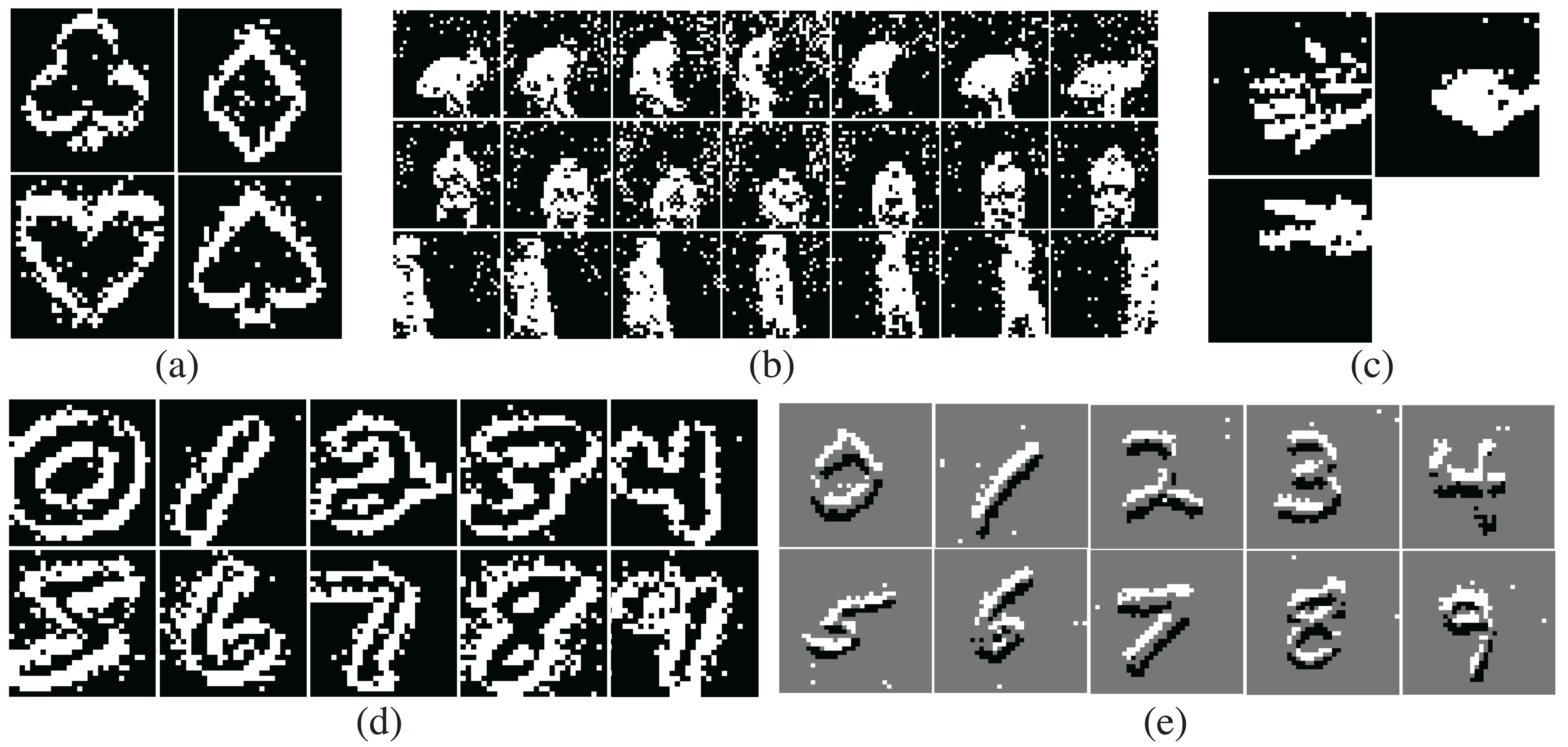}
	\caption{Some reconstructed images from the used datasets. (a): POKER-DVS dataset. (b): AER Posture dataset (rows from top to bottom represent BEND, SITSTAND and WALK respectively). (c): GESTURE-DVS dataset. (d): MNIST-DVS dataset. (e): NMNIST dataset.}
	\label{fig:dvspic}	
\end{figure*}
Since the weights are not restricted in a range, weight normalization \cite{goodhill1994role}, which keeps the sum $L$ of the synaptic weights connected to each learning neuron unchanged, is used to ensure an equal use of the neurons:
\begin{equation}\label{normalization}
\hat{w}_{ij} = \frac{w_{ij}}{\sum\limits_{k=1}^{n_e}w_{kj}}L
\end{equation}
where $w_{ij}$ is the synaptic weights from encoding neuron $i$ to learning neuron $j$, $\hat{w}_{ij}$ is the normalized $w_{ij}$,  $n_e$ is the number of encoding neurons.

\begin{table*}[t]
	\caption{Recognition performance on five datasets.}
	\centering	
	\setlength{\tabcolsep}{3mm}{
		\begin{tabular}{|c|c|c|c|c|c|c|c|}
			\hline
			\hline
			\multirow{2}{*}{Model} & \multirow{2}{*}{POKER-DVS}&\multicolumn{3}{|c|}{MNIST-DVS} &\multirow{2}{*}{NMNIST}&\multirow{2}{*}{AER Posture}&\multirow{2}{*}{GESTURE-DVS} \\ 
			\cline{3-5}
			&&{$100\ ms$}&{$200\ ms$}&{full length}&&&\\
			\hline
			Zhao's \cite{zhao2015feedforward}   & 93.00$\%$ & 76.86$\%$ &82.61$\%$&88.14$\%$&85.60$\%$&  99.48$\%$ &  90.50$\%$ \\ 
			\hline
			BOE \cite{peng2017bag}            	       &   93.00$\%$    &        74.60$\%$ &78.74$\%$ &72.04$\%$ &70.43$\%$ &  98.66$\%$   & 88.97$\%$  \\
			\hline
			HFirst  \cite{orchard2015hfirst}         &   94.00$\%$       	 &   55.77$\%$  & 61.96$\%$& 78.13$\%$ & 71.15$\%$  &    94.48$\%$  & 84.75$\%$  	\\ 
		   \hline
			Our Work	                                       &   	99.00$\%$         &    79.25$\%$   &83.30$\%$	&89.96$\%$& 89.70$\%$ & 99.58$\%$   &  95.75$\%$ \\ \hline
			\hline
		\end{tabular}
	}
	\label{table:result}
\end{table*}
\section{Experimental Results}\label{sec:result}
In this section, we evaluate the performance of our proposed approach on AER datasets and compare our approach with other AER recognition methods.
\subsection{Datasets}
Five different datasets are used in this paper to analyze the performance, i.e., POKER-DVS dataset \cite{perez2013mapping,peng2017bag}, MNIST-DVS dataset \cite{lichtsteiner2008128}, NMNIST dataset \cite{orchard2015converting}, AER Posture dataset \cite{zhao2015feedforward} and GESTURE-DVS dataset. Fig. \ref{fig:dvspic} shows some samples of these five datasets.

\subsubsection{POKER-DVS dataset} It contains 100 samples divided from an event stream of poker card symbols with a spatial resolution of 32 $\times$ 32. It consists of four symbols, i.e., club, diamond, heart and spade. 

\subsubsection{MNIST-DVS dataset} It is obtained with a DVS sensor by recording 10,000 original handwritten images in MNIST moving with slow motion. Due to the motion during the recording of MNIST-DVS dataset, the digit appearances in this dataset have far greater variation than MNIST dataset. Thus, the recognition task of MNIST-DVS is more challenging than that of MNIST.
The full length of each recording is about $2000\ ms$ and the spatial resolution is 28 $\times$ 28. 

\subsubsection{N-MNIST dataset} it is obtained by moving an ATIS camera in front of the original MNIST images. It consists of 60,000 training and 10,000 testing samples. The spatial resolution is  34 $\times$ 34.

\subsubsection{AER Posture dataset} It contains 191 BEND action, 175 SITSTAND action and 118 WALK action with a spatial resolution of 32 $\times$ 32. 

\subsubsection{Gesture-DVS dataset} We collect this dataset to further verify the robustness of our approach. We firstly made a fist above the scope of DVS sensor, and then swung the hand down to deliver a gesture. 
We recorded the events triggered by the hand moving down.
The dataset contains three gestures, i.e., rock (a closed fist), paper (a flat hand), and scissor (a fist with the index finger and middle finger extended, forming a V). Each gesture is delivered 40 times in total and the recording of each time has a duration of $50\ ms$.
The events are captured by the DVS sensor with a resolution of 128 $\times$ 128 pixels and scaled to 32 $\times$ 32 pixels in the data preprocessing.
\subsection{Benchmark Methods}
We compare our approach with other three recently proposed AER recognition methods. The first one was proposed by Zhao et al. \cite{zhao2015feedforward}, which extracts the features through a convolution-based network and performs recognition through a tempotron classifier. Tempotron classifier is a supervised learning rule of SNN which specifies the desired status of firing or not for each neuron.
The second one named BOE was proposed by Peng et al. \cite{peng2017bag}, which uses a probability-based method for feature extraction and a support vector machine with linear kernel as the classifier. The third one named HFirst was proposed by Orchard et al. \cite{orchard2015hfirst}, which employs a spiking hierarchical feature extraction model and a classifier based on spike times. We obtain the source codes of these benchmark methods from their authors. 
\subsection{Experiment Settings}
The experiments are run on a workstation with two Xeon E5 2.1GHz CPUs and 128GB RAM. We use MATLAB to simulate Event Flow Segmentation and MuST, the BRIAN simulator \cite{goodman2009brian}  to implement SNN for recognition. 

We randomly partition the used dataset into two parts for training and testing. 
The result is obtained over multiple runs with different training and testing data partitions. We report the final results with the mean accuracy and standard deviation. For fair comparison, the results of methods listed in TABLE \ref{table:result} are obtained under the same experimental settings. The results of benchmark methods are from the original papers\cite{zhao2015feedforward,peng2017bag}, or (if not in the papers) from the experiments using the code \cite{zhao2015feedforward,peng2017bag,orchard2015hfirst} with our optimization.

The constant parameter settings in our approach are summarized here.  We choose four orientations ($0^{\circ}$, $45^{\circ}$, $90^{\circ}$, $135^{\circ}$) and a range of sizes from $3\times 3$ to $9\times9$ pixels with strides of two pixels for Gabor filters. The detailed settings of Gabor filters are listed in TABLE \ref{table:gabor}. 
\begin{table}[t]
	\caption{The parameters of Gabor filters.}
	\begin{center}		
		\begin{tabular*}{\linewidth}{c@{\extracolsep{\fill}}cccc}
			\hline
			\hline
			scale $s$  & 3 & 5 & 7 & 9 \\
			\hline
			effective width $\sigma$  & 1.2 & 2.0 & 2.8 & 3.6 \\
			wavelength $\lambda$ & 1.5 & 2.5 & 3.5 & 4.6 \\
			\hline
			orientation $\theta$ & \multicolumn{4}{c} {$0^{\circ}$, $45^{\circ}$, $90^{\circ}$, $135^{\circ}$} \\
			aspect ratio $\gamma$ & \multicolumn{4}{c} {0.3} \\
			\hline
			\hline 
		\end{tabular*}
	\end{center}
	\label{table:gabor}
\end{table}
These parameter settings have been proved solid on the task of visual feature capturing, and inherited in many works \cite{zhao2015feedforward,orchard2015hfirst,liu2017fast}.
The time constant of feature response $\tau_{leak}$ is set according to the time length of the symbol in each dataset. The $\tau_{leak}$ for POKER-DVS, MNIST-DVS, NMNIST, AER Posture, and GESTURE-DVS dataset is set to $10 \ ms$, $100 \ ms$, $30\ ms$, $100 \ ms$, and $50 \ ms$, respectively.
The time window $t_w$ and threshold $r_{min}$ in coding function are set as $500 \ ms$ and $0.2$ respectively. The parameters of neuron model in the recognition layer are set as follows: $V_{rest} = -65 \ mV$, $E_{exc} = 0 \ mV$, $E_{inh} = -100 \ mV$, $\tau = 100 \ ms$, $V_{t} = -63.5 \ mV$, $V_{plus} = 0.07 \ mV$, $\tau_{thr} = 1e7 \ ms$. The parameters in STDP are set as follows: $\tau_{apre} = 20 \ ms$, $\tau_{apost} = 30 \ ms$, $\tau_{apost2} = 40 \ ms$, $A^+ = 0.1$, $A^- = 0.001$. The other parameters in recognition layer are set as follows. The inhibitory weight $w_{inh}$ is set as $2.4$ and the delay time $t_d$ is set as $0.3 \ ms$. According to the number of samples, the number of learning neurons for POKER-DVS, MNIST-DVS, NMNIST, AER Posture, and GESTURE-DVS dataset are set as $60$, $700$, $1200$, $600$, and $60$. Due to different spatial resolutions, the parameter $L$ in weight normalization is set as $37.5$ for MNIST-DVS dataset, $54.0$ for NMNIST dataset, and $47.0$ for POKER-DVS dataset, AER Posture dataset, and GESTURE-DVS dataset.


\subsection{Performance on Different AER Datasets}
\subsubsection{On POKER-DVS dataset}
For each category of this dataset, 90\% are randomly selected for training and the others are used for testing. 
We obtain the average performance by repeating the experiments 100 times. 

Our approach gets the recognition accuracy of 99.00\% on average, with a standard deviation of 3.84\%.
TABLE \ref{table:result} shows that our approach outperforms Zhao's method \cite{zhao2015feedforward}, BOE \cite{peng2017bag} and HFirst \cite{orchard2015hfirst} 
by a performance margin of 6.00\%, 6.00\% and 
5.00\% respectively. 

\subsubsection{On MNIST-DVS dataset}
This dataset has 10,000 symbols, 90\% of which are randomly selected for training and the remaining ones are used for testing. The performance is averaged over 10 runs. The experiments are conducted on recordings with the first $100\ ms$, $200\ ms$ and full length (about $2000\ ms$) respectively.

\begin{figure}[b]
	\centering
	\includegraphics[width=0.72\linewidth,bb=5 5 480 397]{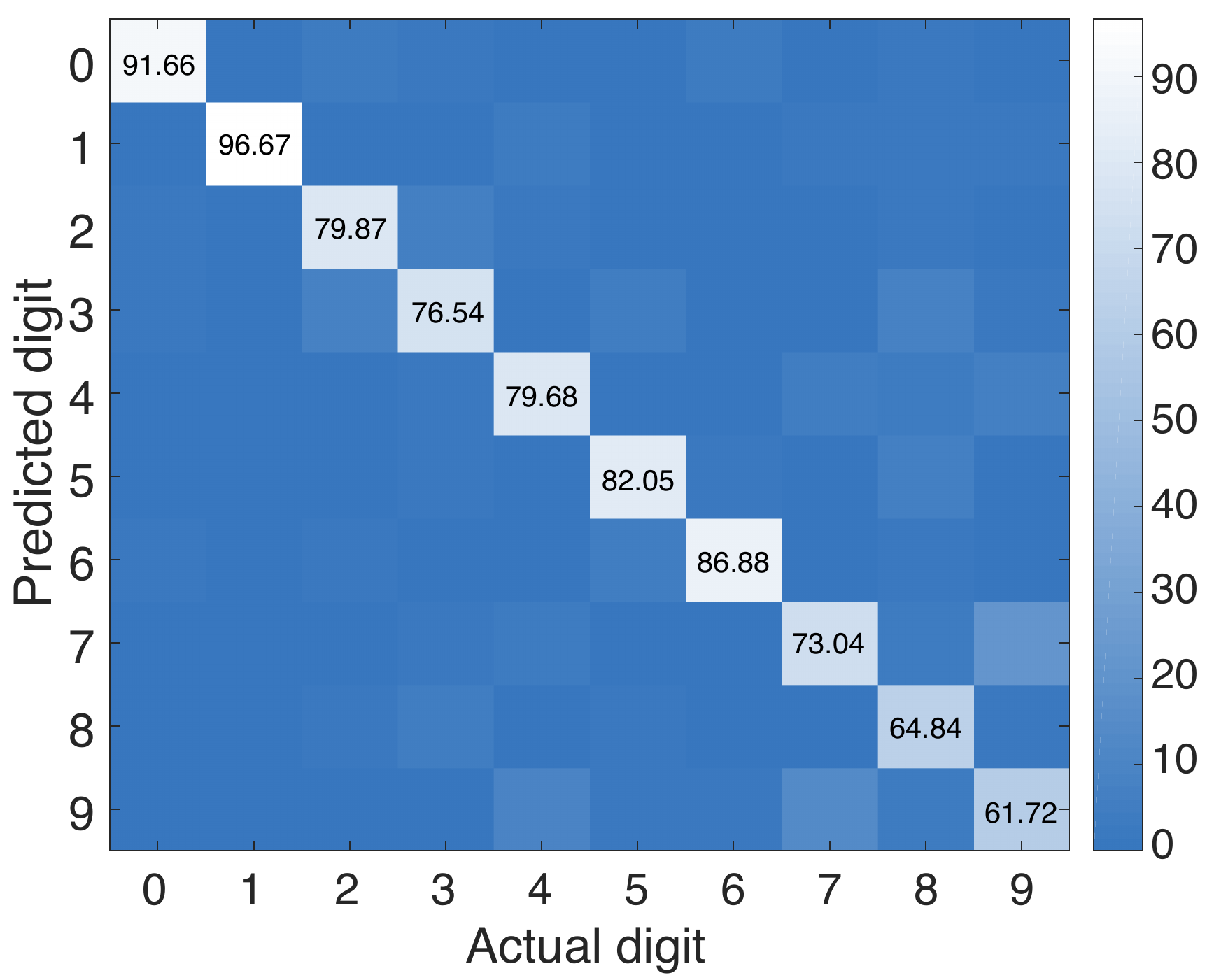}
	\caption{Average confusion matrix of the testing results over 10 runs of MNIST-DVS $100\ ms$ dataset.
	 }
	\label{fig:heatmap}	
\end{figure}

Fig. \ref{fig:heatmap} shows the correct recognition rates  on recordings with $100\ ms$ of each digit along the diagonal and the confusions anywhere else. Digit 1 gets the highest accuracy of 96.67\% because of its simple stroke. 
Confusions occur mostly between digit 7 and 9.
As can be noticed in Fig. \ref{fig:dvspic},  the difference between the two digits is that there is an extra horizontal stroke in 9, which is connected to the above stroke. 
Hence, the learning neurons representing 9 are likely to fire when the input pattern is 7.
Overall, our approach achieves the recognition accuracy of 79.25$\%$, 83.30$\%$ and 89.96$\%$ on the recordings of $100\ ms$, $200\ ms$ and full length.  We can see that our performance becomes better with the longer recordings.
Further, our approach consistently outperforms other methods on recordings with every time length in TABLE \ref{table:result}. 
\subsubsection{On NMNIST dataset}
 This dataset is inherited from MNIST, and has been partitioned into 60,000 training samples and 10,000 testing samples by default. MNIST-DVS and NMNIST datasets are both derived from the original frame-based MNIST dataset. Compared with MNIST-DVS dataset recorded by moving the MNIST images with slow motion, NMNIST dataset is captured by moving the AER sensor. The obtained event streams in two datasets are not the same. 
 
 Our approach gets the recognition accuracy of 89.70\%.
 TABLE \ref{table:result} shows that the recognition performance of our approach is higher than that of Zhao's method \cite{zhao2015feedforward}, BOE \cite{peng2017bag} and HFirst \cite{orchard2015hfirst}. In addition, compared with Iyer \& Basu's unsupervised model \cite{iyer2017unsupervised} on NMNIST, which achieves the accuracy of 80.63\%, our approach can give higher accuracy of 89.70\%.
 
\subsubsection{On AER Posture dataset}
In this dataset, we randomly select 80\% of human actions for training and the others for testing. 
The experiments are repeated 10 times to obtain the average performance.
The results are listed in TABLE \ref{table:result}. 

 The recognition accuracy obtained by our approach is 99.58\%. Our approach has a performance that is comparable to Zhao's \cite{zhao2015feedforward}, higher than BOE \cite{peng2017bag} and HFirst \cite{orchard2015hfirst}.
\subsubsection{On Gesture DVS dataset}
This dataset has 3 categories, each with 40 samples. For each category, 90\% samples are randomly selected for training and the others are for testing. We perform the experiments 100 times and average the performance.
In this dataset, the position of the hand in each sample is not constant and some portion of the player's forearm is sometimes recorded. These randomnesses increase the difficulties of the recognition task of this dataset.

Our approach achieves the recognition accuracy of 95.75\%, with a standard deviation of 5.49\%. TABLE \ref{table:result} shows that our approach outperforms Zhao's method \cite{zhao2015feedforward}, BOE \cite{peng2017bag} and HFirst \cite{orchard2015hfirst} 
by a performance margin of 5.25\%, 6.78\% and 
11.00\% respectively. 
\subsection{Analyses of the MuST}
In this section, we carry out experiments to analyze the effects of MuST from two aspects: the temporal coding function and the spatial fusion method. The experiments are conducted on POKER-DVS dataset, AER Posture dataset , 1,000 samples of MNIST-DVS $100ms$ dataset and GESTURE-DVS dataset. For each dataset, the experiment settings are the same as the previous section.
\subsubsection{Effects of the temporal coding function}
\begin{table}[h!]
	\caption{Accuracy with linear coding function and natural logarithm coding function.}
	\begin{center}		
		\begin{tabular*}{\linewidth}{c@{\extracolsep{\fill}}cccc}
			\hline
			\hline 
			Method  & POKER & Posture &MNIST &GESTURE\\
			\hline
			Linear Coding & 95.25$\%$  &96.69$\%$ & 73.30$\%$&95.25$\%$\\
			\textbf{Log Coding} & \textbf{99.00}$\textbf{\%}$& \textbf{99.58}$\textbf{\%}$&
			\textbf{76.90}$\textbf{\%}$&\textbf{95.75$\%$}\\
			\hline
			\hline 
		\end{tabular*}
	\end{center}
	\label{table:codingAcc2}
\end{table}
We compare the performance of the approach using conventional linear coding function \cite{zhao2015feedforward,liu2017fast} and the proposed natural logarithm coding function. The linear coding function is set as follows: $t_{spike}=-ar+b$, where $a=t_w/r_{max}$ and $b = t_w$. As shown in TABLE \ref{table:codingAcc2} \footnote{Due to the space limit, names of the dataset are abbreviated accordingly.}, 
the proposed logarithm coding function achieves higher performance than the linear one on three datasets. 

\begin{figure}[t]
	\centering
	\includegraphics[width=0.72\linewidth,bb=20 12 390 310]{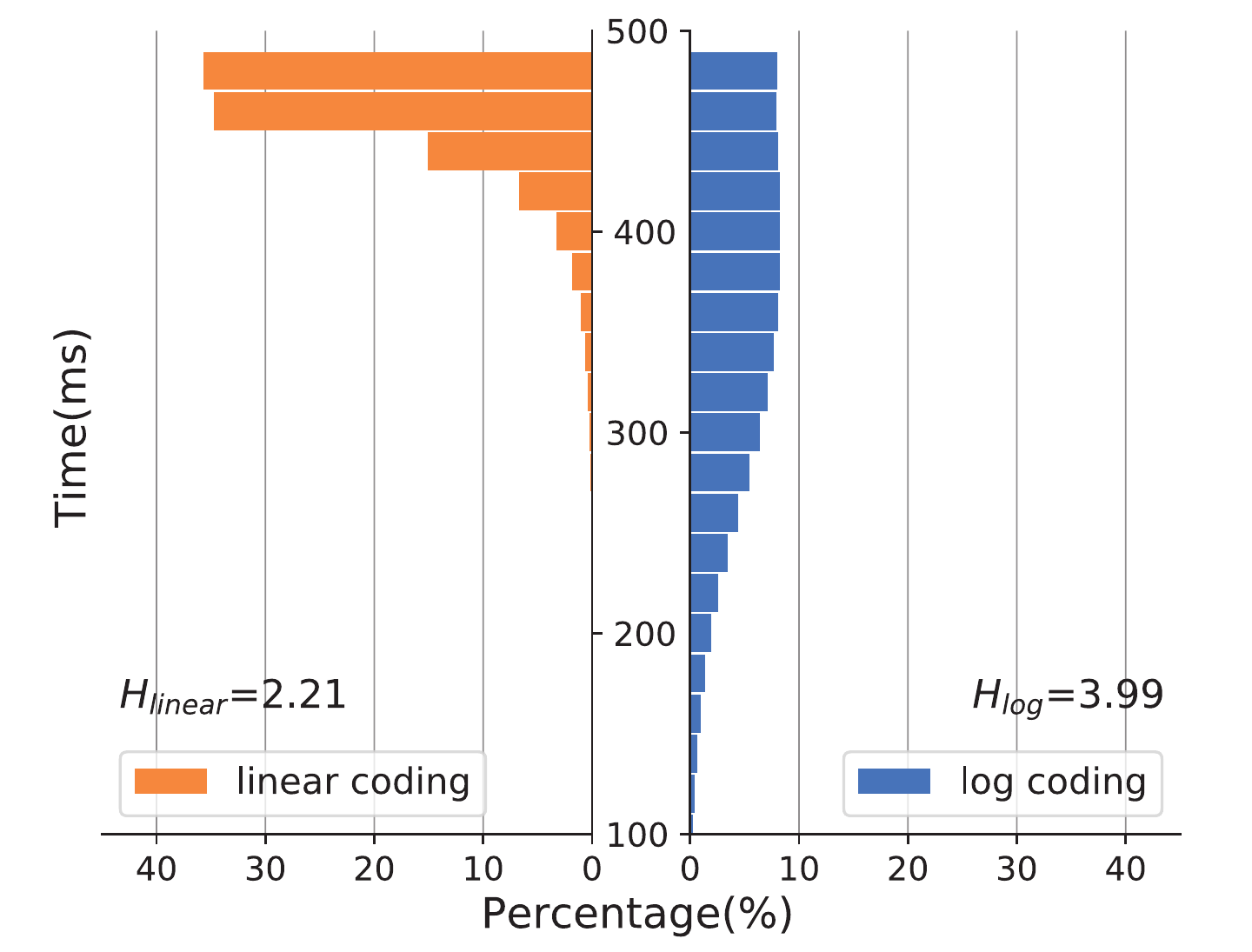}
	\caption{Spike timing distributions with linear coding function and natural logarithm coding function on MNIST-DVS $100ms$ dataset. Each bin has a nonoverlapping temporal span of $20\ ms$, and the time window $t_w$ is $500\ ms$. The height of each bin indicates the average proportion of the spike timings in the corresponding time span. 
		The natural logarithm coding function evens the distribution of the timings of feature spikes.
		 The information entropy $H_{linear}$ and $H_{log}$ are $2.21$ and $3.99$, respectively.
	}
	\label{fig:codingfunction}	
\end{figure}

As reported in Fig. \ref{fig:codingfunction}, using linear coding function, feature spikes which are emitted early has a quite sparse temporal distribution. For example, the linear coding function generates only approximately $8\%$ spikes before $400\ ms$. The feature spikes with sparse distribution are hard to accumulate potential of the learning neurons high enough to emit spikes.  Thus, the information in these feature spikes cannot be transmitted to the learning neurons. 
As the temporal distribution of spikes becomes denser, potential of the learning neuron becomes higher and emits more spikes gradually.
 According to the STDP learning, the synaptic weight is updated when there is a presynaptic spike or a postsynaptic spike. Therefore,  the learning and the recognition is mostly affected by the feature spikes emitted later. 
As can be seen in Fig. \ref{fig:codingfunction}, the proposed logarithm coding function evens the temporal distribution of spikes, so that the features can be used equally to a large extent. 
The analysis can also be given in another aspect.
Considering that the information in SNN is represented by the spike timings, we evaluate the information carried by the feature spikes generated by these two coding functions using information entropy of the spikes. The information entropy with the higher value means the corresponding feature representation contains more information of the features.
The information entropy is calculated as:
\begin{equation}\label{entropy}
H = -{\sum\limits_{i}p_i}\log_2 p_i
\end{equation}
where $p_i$ denotes the portion of spikes located within the $i$-th temporal bin. The information entropy $H_{log}$ of feature spikes generated by natural logarithm coding function is  $3.99$, which is higher than $H_{linear}$ generated by linear coding function with $2.21$.
This suggests that the obtained MuST feature representation is more informative and the proposed natural logarithm coding function conveys more information of the features into the spikes, which contributes to the recognition using SNN.
\subsubsection{Effects of the spatial fusion method}  
In our approach, the spatial features of AER events are extracted from two aspects, i.e., scales and orientations. There exist four spatial fusion options for features, i.e., multiscale fusion, multi-orientation fusion, no fusion and full fusion. 
We will compare our approach with those using full fusion, multi-orientation fusion  and no fusion instead of multiscale fusion to provide the analyses. 
\textbf{Multiscale fusion} fuses features of multiple scales having same orientation $\theta$ and position $(x,y)$ in feature maps into a spike-train for an encoding neuron, which is comprised of a set of $t_{spike}$ in Equation (\ref{equation:fusion}) where  $S =\left\{3,5,7,9 \right\}$ and $\Theta =\left\{ \theta \right\}$.
\textbf{Multi-orientation fusion} fuses features of multiple orientations having same scale $s$ and position $(x,y)$ into a spike-train, which is comprised of $t_{spike}$ in Equation (\ref{equation:fusion}) where  $S =\left\{ s \right\}$ and $\Theta =\left\{ 0^{\circ}, 45^{\circ}, 90^{\circ}, 135^{\circ} \right\}$. \textbf{No fusion} does not fuse any feature spike, and the feature spike having scale $s$, orientation $\theta$ and position $(x,y)$  can be expressed using Equation (\ref{equation:fusion}) where $S =\left\{ s \right\}$ and $\Theta =\left\{ \theta \right\}$.
\textbf{Full fusion} fuses all feature spikes having the same position $(x,y)$  into a spike-train, that is comprised of $t_{spike}$ in Equation (\ref{equation:fusion}) where  $S =\left\{ 3,5,7,9 \right\}$ and $\Theta =\left\{ 0^{\circ}, 45^{\circ}, 90^{\circ}, 135^{\circ} \right\}$.
TABLE \ref{table:fusionAcc2} reports the recognition accuracy and the required number of parameters of these four methods.
We will give the analyses via 3 comparisons: 
\begin{table}[t]
	\caption{Accuracy and required parameters with four fusion methods.}
	\begin{center}  
		\begin{tabular*}{\linewidth}{c@{\extracolsep{\fill}}clrr}
			\hline
			\hline 
			\textbf{Dataset}  & & \textbf{Accuracy}&\textbf{Params }& \\
			\hline
			POKER-DVS&&&\\
			\multicolumn{2}{l}{\qquad \textbf{Multiscale Fusion}}   &
			\textbf{99.00}$\textbf{\%}$& \textbf{0.43M}\\
			\multicolumn{2}{l}{\qquad Multi-Orientation Fusion} & {94.50$\%$} &  {0.43M}\\
			\multicolumn{2}{l}{\qquad No Fusion} &96.63$\%$ & 1.72M \\
			\multicolumn{2}{l}{\qquad Full Fusion} & 85.50$\%$&  0.11M \\
			\hline
			AER Posture&&&\\
			\multicolumn{2}{l}{\qquad \textbf{Multiscale Fusion}}   &
			\textbf{99.58}$\textbf{\%}$
			&\textbf{{4.67M}}\\
			\multicolumn{2}{l}{\qquad Multi-Orientation Fusion} &99.00$\%$ & {4.67M}\\
			\multicolumn{2}{l}{\qquad No Fusion} &92.17$\%$ & 17.57M \\
			\multicolumn{2}{l}{\qquad Full Fusion} & 90.56$\%$&  1.44M \\
			\hline
			MNIST-DVS&&&\\
			\multicolumn{2}{l}{\qquad \textbf{Multiscale Fusion}}   &
			\textbf{76.90}$\textbf{\%}$ & {\textbf{4.34M}}\\
			\multicolumn{2}{l}{\qquad Multi-Orientation Fusion} & 57.97$\%$& {4.34M}\\
			\multicolumn{2}{l}{\qquad No Fusion} &75.62$\%$ & 15.87M \\
			\multicolumn{2}{l}{\qquad Full Fusion} & 54.64$\%$&  1.46M \\
			\hline
			GESTURE-DVS&&&\\
			\multicolumn{2}{l}{\qquad \textbf{Multiscale Fusion}}   &
			\textbf{95.75}$\textbf{\%}$ & {\textbf{0.43M}}\\
			\multicolumn{2}{l}{\qquad Multi-Orientation Fusion} & 90.83$\%$& {0.43M}\\
			\multicolumn{2}{l}{\qquad No Fusion} &73.25$\%$ & 1.72M \\
			\multicolumn{2}{l}{\qquad Full Fusion} & 80.58$\%$&  0.11M \\
			\hline
			\hline
		\end{tabular*}
	\end{center}
	\label{table:fusionAcc2}
\end{table}

First, both multi-orientation fusion and multiscale fusion fuse the features along their corresponding aspect, and require the same number of parameters since the number of scales and orientations are the same in our settings. 
But multi-orientation fusion yields a lower performance, as shown in TABLE \ref{table:fusionAcc2}.  An important factor to affect the result of these two fusion methods is the correlation among data sources. A high correlation between features implies features contain similar information, while a lower feature correlation means that features have richer diversity. 
It is expected that highly correlated features are fused together to one neuron, while low correlation features are separated to different neurons, so that learning neurons can distinguish various patterns of the fused spikes more easily.

\begin{figure}[t]
	\centering
	\includegraphics[width=0.8\linewidth,bb=20 10 395 320]{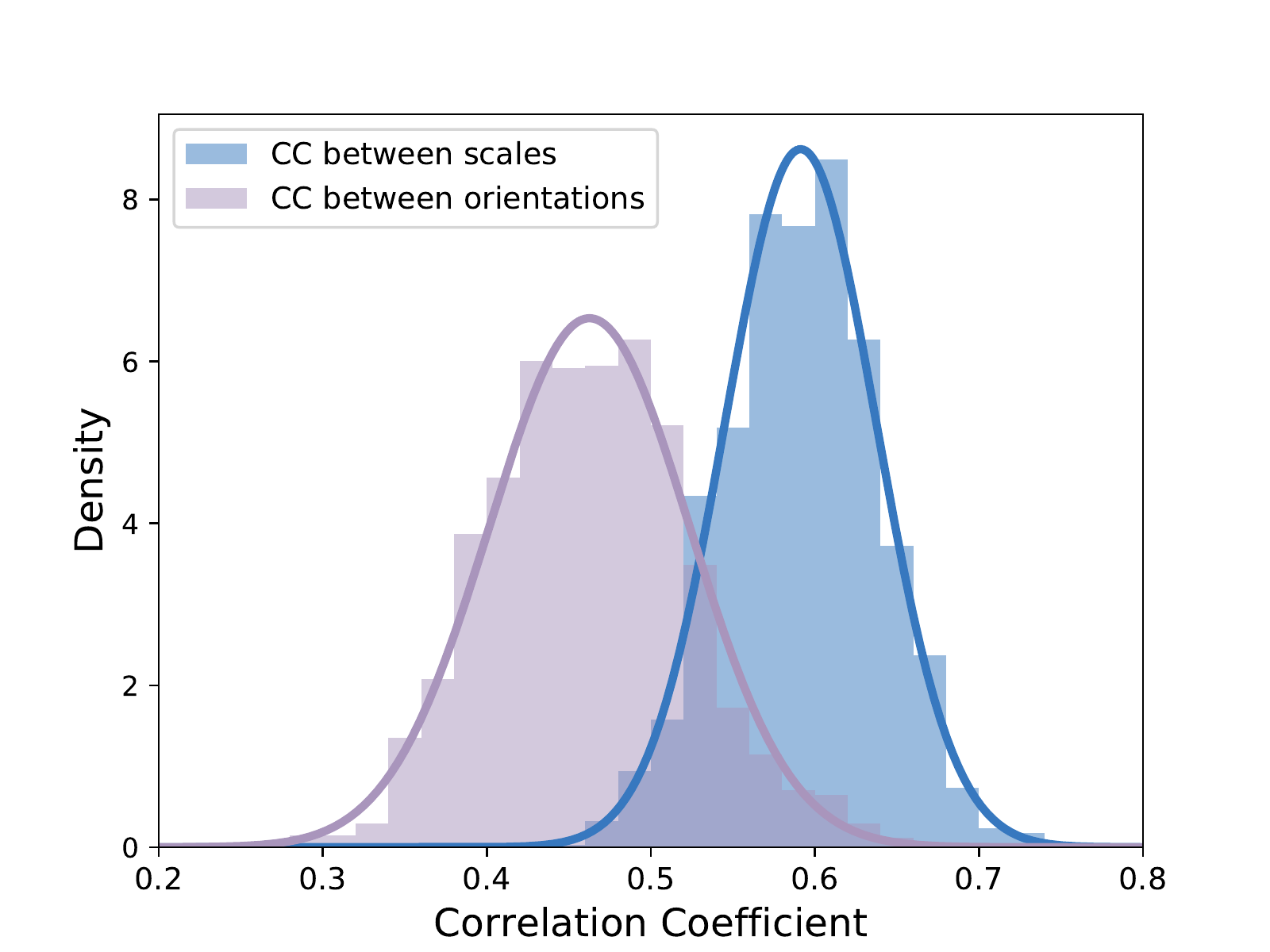}
	\caption{The normalized histograms of correlation coefficient (CC),  and their fitted probability density functions.
		 Each CC value is derived from a pair of response series of different scales or of different orientations, and the distribution consists of CCs for every possible pair. 
		 Each bin has a nonoverlapping span of $0.02$, and the height of each bin indicates the density of CC values in the corresponding span. 
		CC between orientations is on average lower than CC between scales. 
	}
	\label{fig:cc}
\end{figure}
We use the correlation coefficient (CC) to measure the correlation and randomly choose 1000 samples of MNIST-DVS $100ms$ dataset for illustration.
For the $i$-th sample, CC between scales is obtained by averaging the Pearson CCs of pairwise scale maps having the same orientations:
\begin{equation}\label{cc}
CC_{s}^i =\frac{ {\sum\limits_{\theta = 1}^{n_{\theta}}}{\sum\limits_{s = 1}^{n_s}}{\sum\limits_{s' = s+1}^{n_s}} \rho(r(s,\theta),r(s',\theta))}{M}
\end{equation}
where $r(s,\theta)$ represents the vector of $C1$ responses at scale $s$ and orientation $\theta$, $\rho(A,B)$ denotes Pearson correlation coefficient between vector A and vector B, $M = n_{\theta}\tbinom{2}{n_s}$ denotes the number of pairs of feature vectors. 
CC between orientations is obtained in the same way but with pairwise orientation maps having the same scales. We can see from Fig. \ref{fig:cc}, CC between orientations is on average lower than CC between scales. 
Specifically, there are only about 3\% of values of CC between scales less than $0.5$, but approximately 74\% of CC between orientations less than 0.5.
It demonstrates that features of different orientations have lower correlation than those of different scales. 
Multi-orientation fusion brings together diverse information into one neuron to express, and separates similar information to different neurons to express. Therefore, the recognition network are hard to learn the spike patterns, which results in a lower performance.

Second, we notice that the method without fusion maintains relatively high accuracies on three datasets but requires larger number of parameters in recognition part. 
Without fusion, each encoding neuron represents a specific spatio-temporal feature.
As shown in TABLE \ref{table:fusionAcc2}, this method will require larger computation resource. 
Nevertheless, multiscale fusion can achieve a competitive result with 
more efficient resource usage, which is well suited for resource-constrained neuromorphic devices.


Third, full fusion fuses all the spatio-temporal features of a position and obtains the worst result on three datasets. The fusion degree of full fusion is higher than other three fusion methods. Although it requires least computation resource, this method faces severe limitation of feature expression and therefore has a poor recognition accuracy.
\section{Conclusion}\label{sec:conclu}

In this paper, we propose an unsupervised recognition approach for AER object. The proposed approach presents a MuST representation for encoding AER events and employs STDP for object recognition with MuST. 
MuST exploits the spatio-temporal information encapsulated in the AER events and forms a feature representation that contributes to the latter recognition.
Experimental results show the effects of MuST from both temporal and spatial perspectives.
MuST, with even temporal distribution, has been shown informative and can improve the performance of recognition.
MuST also fuses highly correlated features, forming a compact spike representation, which consumes less computational resource while still maintaining comparable performance. The recognition process employs a SNN trained by the triplet STDP, which does not require a teaching signal or setting the desired status of neurons.
Compared with other state-of-the-art supervised benchmark methods, our approach yields comparable or even better performance on five AER datasets, including a new dataset named GESTURE-DVS that further verifies the robustness of our approach.

\ifCLASSOPTIONcaptionsoff
  \newpage
\fi



%
%
%

\bibliographystyle{IEEEtran}

\bibliography{cite}

\end{document}